\documentclass[10pt,twocolumn,letterpaper]{article} 

\usepackage{avss}
\usepackage{times}
\usepackage{epsfig}
\usepackage{graphicx}
\usepackage{amsmath}
\usepackage{amssymb}
\usepackage[squaren,Gray]{SIunits}
\usepackage{multirow} 
\usepackage[table,svgnames]{xcolor}
\usepackage{hhline}
\usepackage{rotating}
\usepackage{tabularx}
\usepackage{url}
\usepackage{setspace}
\usepackage{picins}
\usepackage{listings}
\usepackage[vlined,ruled]{algorithm2e}

\makeatletter
\def\hlinewd#1{%
\noalign{\ifnum0=`}\fi\hrule \@height #1 %
\futurelet\reserved@a\@xhline}
\makeatother

\avssfinalcopy 

\def\avssPaperID{41} 

\ifavssfinal\fi

\begin{document}

\SetKwInOut{Input}{input}\SetKwInOut{Output}{output}

\title{A generic framework for video understanding\\
applied to group behavior recognition}

\author{Sofia Zaidenberg, Bernard Boulay,  Fran\c cois Br\'emond\\
Inria\\
STARS team \\
2004 Route des Lucioles - BP 93\\
06902 Sophia Antipolis (France)\\
{\tt\small firstname.name@inria.fr}
}

\maketitle

\begin{abstract}
This paper presents an approach to detect and track groups of people in
video-surveillance applications, and to automatically recognize their behavior.
This method keeps track of individuals moving together by maintaining a spacial
and temporal group coherence. First, people are individually detected and 
tracked. Second, their trajectories are analyzed over a temporal window and
clustered using the Mean-Shift algorithm. 
A coherence value describes how well a set of people can be described as a
group. Furthermore, we propose a formal event description language. The group
events recognition approach is successfully validated on 4 camera views from 3
datasets: an airport, a subway, a shopping center corridor and an entrance hall.
\end{abstract}

\section{Introduction}

\begin{figure}[h]
  \centering
    \includegraphics[width=\linewidth]{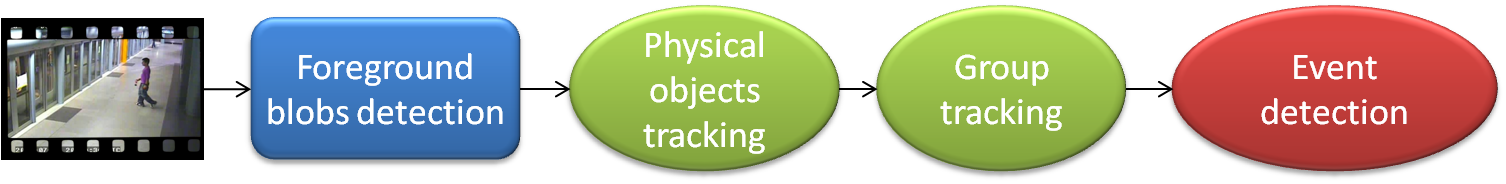}
 \caption{Description of the proposed video understanding system}
  \label{fig:subgraph}
\end{figure}

In the framework of a video understanding system (figure~\ref{fig:subgraph}),
video sequences are \emph{abstracted} in physical objects: objects of interest
for a given application. Then the physical objects are used to recognize events.
In this paper, we are interested by the group behavior in public spaces. Given a
set of detected and tracked people, our task is finding associations of those
people into spatially and temporally coherent groups, and detecting events
describing group behavior.

\addtolength{\textfloatsep}{-5mm}

\begin{figure}
 \centering
 \includegraphics[width=0.49\linewidth]{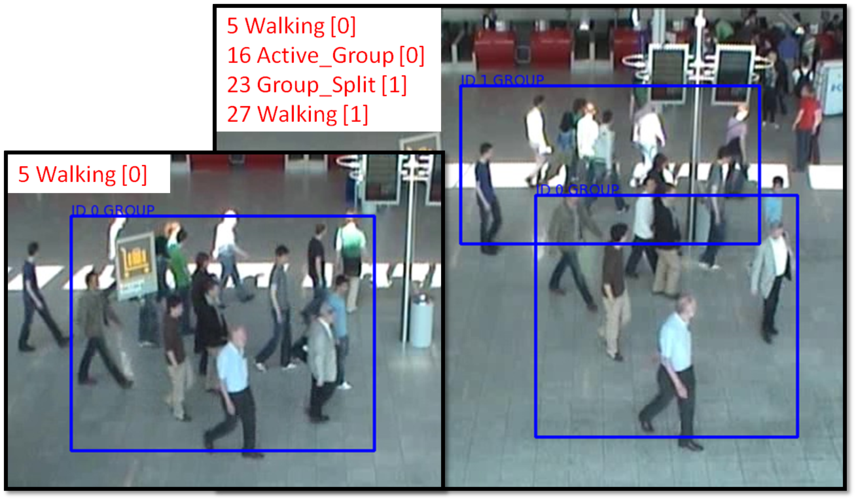}
 \includegraphics[width=0.49\linewidth]{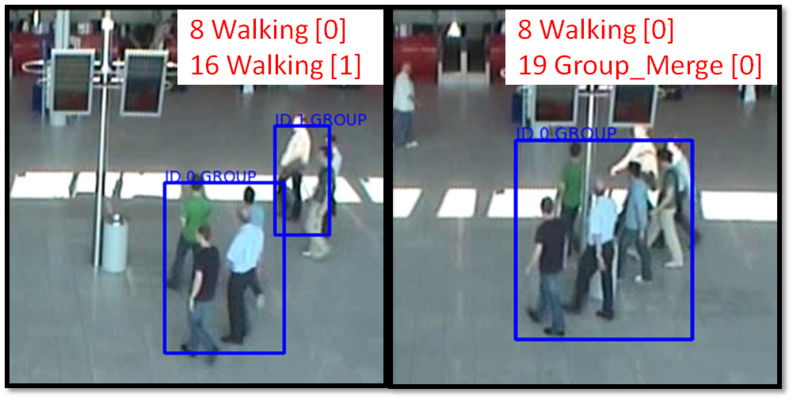}
 \caption{Group event recognition on Eindhoven airport sequences. Left: The
group is detected as splitting into 2 sub-groups. Right: Two groups are detected
as merging into one.}
 \label{fig:eindhoven}
\end{figure}

Tracking people, and especially groups of people in relatively an unconstrained,
cluttered environment is a challenging task for various reasons.
In~\cite{Ge09}, Ge \emph{et al.} propose a method to
discover small groups of people in a crowd based on a bottom-up hierarchical
clustering approach. Trajectories of pedestrians are clustered into groups based
on their closeness in terms of distance and velocity. The experiments of this
work have been made on videos taken from a very elevated viewpoint, providing
few occlusions. Haritaoglu \emph{et al.}~\cite{Haritaoglu2001} detect and track
groups of people as they shop in a store. Their method is based on searching
strongly connected components in a graph created from trajectories of individual
people, following the idea that people belonging to the same group have a lower
inter-distance. This method however does not allow group members to move away
and return to the group without being disconnected from it. Furthermore, the
application of a shopping queue lacks genericity (people are rather static and
have a structured behavior), it is not clear how well this method is adaptable
to another context of use. Other approaches, such as~\cite{Naturel2008}, aim at
detecting specific group-related events (\emph{e.g.} queues at vending machines)
without tracking. Here again, the method does not aim at consistently tracking a
group as its dynamics vary. In~\cite{Jacques2007}, an algorithm for group
detection and classification as voluntary or involuntary (\emph{e.g.} assembled
randomly due to lack of space) is proposed. A top-down camera is used to track
individuals, and Voronoi diagrams are used to quantify the sociological concept
of personal space. No occlusion handling is done in this work hence the
applicability to other points of view of the camera or to denser scenes is
questionable. Figure~\ref{fig:eindhoven} shows the result of our event
recognition on tracked groups on a dataset recorded at the Eindhoven airport.

Event recognition is a key task in automatic understanding of video sequences.
In this work we are mainly interested in group events, but the usual techniques
can be applied to different kinds of objects (person, vehicle, group,...). The
typical detection algorithm (figure~\ref{fig:subgraph}) takes as input a video
sequence and extracts interesting objects (physical objects). This
\textbf{abstraction stage} is the layer between the image and the semantic
worlds. Then, these objects of interest are used to \textbf{model} events.
Finally, the events are \textbf{recognized}. The abstraction stage determines
which modeling techniques can be applied.\\
The possible abstraction technique can be pixel based~\cite{Bobick01} or object
based~\cite{Vu03}. The first kind of techniques is not well adapted for groups.
Indeed, persons belonging to the same group are not necessary physically linked.
With object abstraction, a video sequence is represented thanks to the detected
objects (persons, vehicles,...) and their associated properties (speed,
trajectory,...). In the literature, several approaches use this abstraction
level~\cite{Romdhane2010}, because an activity can be naturally modeled based on
those properties.

Lavee \emph{et al.}~\cite{Lavee09} classify existing event modeling techniques
in three categories: the pattern recognition models, the state based models and
the semantic models.\\
The \textbf{pattern recognition models} are classical recognition techniques
using classifiers as the nearest neighbor method, boost techniques, support
vector machines and neural networks~\cite{Chen06}. These techniques are well
formalized. But adding new types of events implies the training of new
classifiers.\\
The \textbf{state based models} formalize the events in spatial and temporal
terms using semantic knowledge: Finite State Machines (FSM), Bayesian Networks
(BN), Hidden Markov Models (HMM), Dynamic Bayesian Networks (DBN) and
Conditional Random Fields (CRF). The HMMs and all their variants are heavily
used for event modeling~\cite{Duong05}. They take the advantages of the FSM
(temporal modeling) and of the BNs (probabilistic modeling). But due to the
nature of the HMMs (time sliced structure), the complex temporal relations
(\emph{e.g.} during) are not easily modeled. Lin \emph{et al.}~\cite{Lin10}
propose an asynchronous HMM to recognize group events. Brdiczka \emph{et al.}
construct in~\cite{Brdiczka05} HMMs upon conversational hypotheses to model
group events during a meeting. One drawback of the modified HMM methods is that
since the classical structure of HMMs is modified, efficient algorithms can not
be applied without approximation.\\
The \textbf{semantic models} define spatio-temporal relations between sub-events
to model complex events. Due to the nature of these models, the events must be
defined by an expert of the application domain. Moreover, these models are often
deterministic. Several techniques are studied: grammar based models, Petri nets
(PN), constraint solving models and logic based models.\\
As shown in this section, the quantity of techniques for abstraction and event
modeling is huge. In this paper, we propose a framework (\emph{ScReK}: Scenario
Recognition based on Knowledge) to easily model the semantic knowledge of the considered
application domains: the objects of interest and the scenario models, and to
recognize events associated to the detected group based on spatio-temporal
constraints.\\

In the rest of this paper, we first describe our technique to detect and track
groups of people (section~\ref{sec:groupTracking}), then we describe our event
detection method applied to tracked groups (section~\ref{sec:events}). Section~\ref{sec:res} presents evaluations.

\section{Group Tracking}
\label{sec:groupTracking}

Given a set of detected and tracked people, the proposed method focuses mainly
on the task of finding associations of those people into spatially and
temporally coherent groups. The human definition of a group is \emph{people
that know each other or interact with each other}. In fact,
according to McPhail~\cite{McPhail82}: \emph{Two defining criteria of a group
[are] \emph{proximity} and/or \emph{conversation} between two or more persons}.
It is quite difficult to directly detect people interactions and conversation in
a video or the fact that people know each other. For automatic recognition we
derive this definition: \emph{two or more people who are spatially and
temporally close to each other and have similar direction and speed of
movement}, or better: \emph{people having similar trajectories}.


Group tracking is based on people detection. 
The people detection can be performed by various methods. 
We have compared several methods and chosen the best one, it is based on
background-subtraction described in~\cite{Yao2007} because of the quality of
its results (see table~\ref{tb:resPeopleDetectionFiltered} for a comparison
of several methods).

Blobs of foreground pixels are grouped to form physical objects (also called
\emph{mobiles}) classified into predefined categories based on the 3D size of
objects (using a calibrated camera): {\small \verb+GROUP_OF_PERSONS+}, {\small
\verb+PERSON+} and {\small\verb+NOISE+}. When people overlap (which happens
quite often with a low viewpoint, such as in figure~\ref{fig:vanaheim}) or are
too close to each other, segmentation fails to split them and they are detected
as a single object classified as {\small \verb+GROUP_OF_PERSONS+} because its
size is bigger than the size of a single person. Thoses classes of objects are
specified using gaussian functions. Mean, sigma, min and max values are provided
for each class and a score is computed representing how well an object's
dimensions fit in each category. The category with the best score is assigned as
the class of the object. Detected objects at each frame are tracked consistently
on the long term using a multiple feature-based tracker~\cite{Chau2011REAL}.

\begin{figure}
 \centering
 \includegraphics[width=0.49\linewidth]{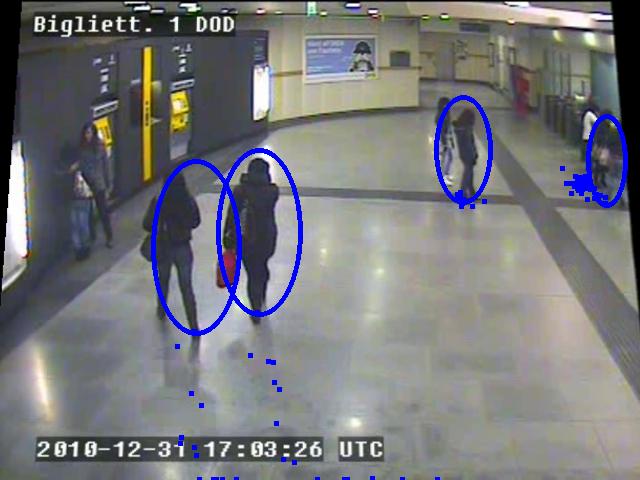}
 \includegraphics[width=0.49\linewidth]{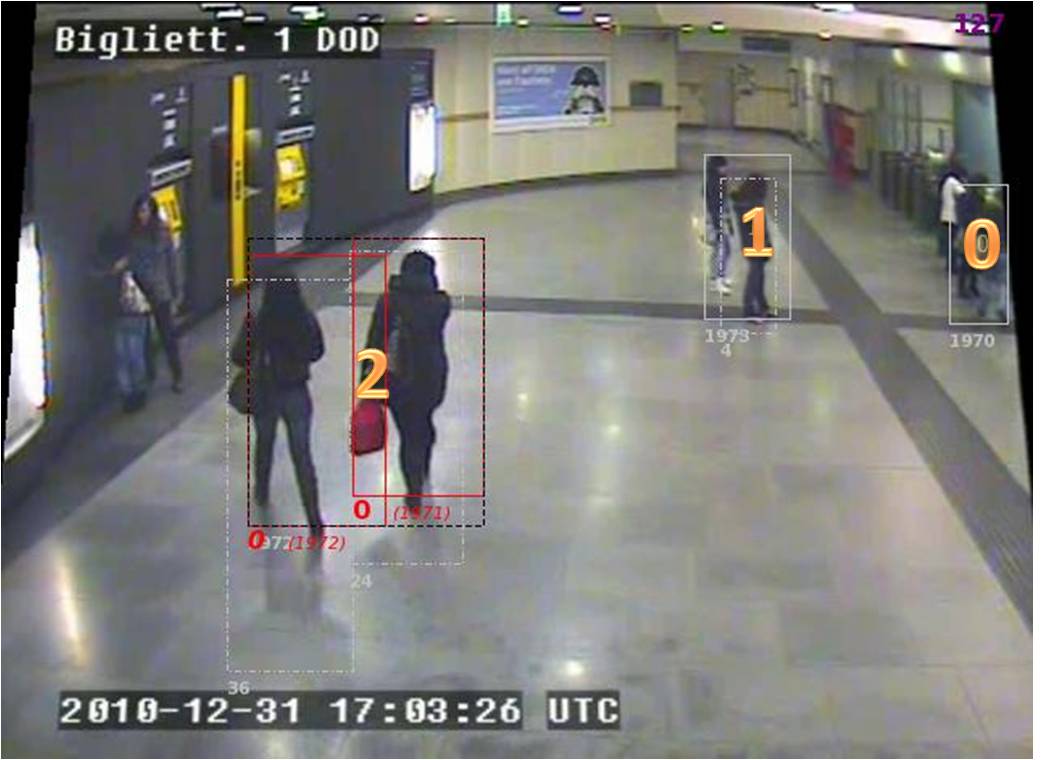}
 \caption{Example of 5 tracked people clustered into 3 trajectory clusters. A
group is created from cluster 2.}
 \label{fig:actorCluster}
\end{figure}

Individual trajectories are the input of the group tracking algorithm, which is
divided into four parts: creation, update, split/merge and termination.
In order to detect temporally coherent groups, we observe people trajectories
over a \emph{time window}, denoted delay $T$. In the experiments presented
section~\ref{sec:res}, we used $T=20$ frames. Working at frame $t_c-T$, $t_c$
being the current frame of the video stream, we cluster trajectories of
individuals between frames $t_c-T$ and $t_c$ to find similar trajectories,
representative of groups. We choose the Mean-Shift clustering
algorithm~\cite{meanshift} because it does not require to set as input the
number of clusters. However, Mean-Shift does require a \emph{tolerance}
parameter determining the size of the neighborhood for creating clusters.
Figure~\ref{fig:actorCluster} shows the input prints and the clustering result.

A trajectory is defined as $Traj=\{(x_i,y_i), i=0 \dotso T-1\} \cup
\{(s_{x_i},s_{y_i}), i=1 \dotso T-1\}$ where $(x_i,y_i), i \in [0;T-1]$ in each
trajectory is the position of a group in the same frame $i$, and
$(s_{x_i},s_{y_i})=speed(i-1,i), i \in [1;T-1]$ is the speed of the group
between frames $i-1$ and $i$. If $k$ positions on the trajectory are missing
because of lacking detections, we interpolate the $k$ missing positions between
known ones. Each trajectory is a point in a $2(2T-1)$-dimensional space.
Mean-Shift is applied on a set of such points.
To make the approach more generic and being able to add other features, we
normalize the values using minimum and maximum ranges. The range of positions on
the ground plane is determined by the field of view. The minimum speed is 0 and
the maximum speed is set to $\unit{10} \meter\per\second$, greatly exceeding all
observed values. From the raw value of $x$, $y$ and $s$ (the speed) denoted by
$r \in [min,max]$, we compute the corresponding normalized value $n \in [0,1]$
as: $n=\frac{r-min}{max-min}$, where $min$ and $max$ are the respective minimum
and maximum values. 
We set the \emph{tolerance} to $0.1$, considering grouping trajectories distant
by less than 10\% of the maximum. This value is quite low because clustering is
used only to group very close people, the case where people temporarily split
being handled by the update step described below.

We characterize a group by three properties: the average over the frames in
which the group is detected of the inter-mobile distance and the average over
frames of standard deviations of speed and direction. These properties enable
the definition of a coherence criterion:
$groupIncoherence=\omega_{1}\cdot distanceAvg+\omega_{2}\cdot
speedStdDev+\omega_{3}\cdot directionStdDev$, where the weights $\omega_{1}$,
$\omega_{2}$ and $\omega_{3}$ are normalization parameters. We use
$\omega_{1}=7$ and $\omega_{2}=\omega_{3}=5$ to slightly favor distance over
speed and direction similarity which are quite noisy. With this definition, a
low value of $groupIncoherence$ is significative of a group.

Groups are created from clusters of more than one physical object.
In the case where one {\small \verb+GROUP_OF_PERSONS+} object is detected at
frame $t_c-T$, we analyze its trajectory through the time window. If this object
stays the size of a group, or is close to other objects, we can create a group
and compute its $groupIncoherence$. If the resulting value is low enough, we
keep the created group. In case of a single {\small \verb+GROUP_OF_PERSONS+}
object, the $groupIncoherence$ value is naturally very low because of a null
$distanceAvg$ component. The \textbf{creation} step is made up of these two
cases.

Group dynamics vary. Sometimes all group members do not have similar
trajectories, for example when the group is waiting while one member buys a
ticket at a vending machine. Clustering is not enough to correctly
\textbf{update} an existing group in that case.
First, we try to associate clusters with groups existing at the previous frame,
using the notion of \emph{probable group} of a mobile, defined hereafter. During
tracking, mobiles detected at different frames are connected by probabilistic
links in order to track consistently the same real objects. We use the term
\emph{father} and \emph{son} for the mobiles resp. in the oldest and most recent
frame of the link. If a father, within a window of $T$ frames, of the mobile $m$
was in a group $g$ and the link probability between father and son is above a
given threshold (a value of $0.6$ is usually used in the experiments
section~\ref{sec:res}), then the father's group $g$ is called the probable group
of the mobile $m$: $PG(m)=g$. Each cluster $c$ is associated with the probable
group of most mobiles in the cluster: $G(c)=\operatorname{argmax}_{g \in
\{g_i^c\}}|\{g_i^c|g_i^c=g\}|$, where $G(c)$ is the group associated with
cluster $c$ and $\{g_i^c\}=\{PG(m_i^c)\}$ the set of probable groups of mobiles
belonging to cluster $c$ ($\{m_i^c\}$ being the set of mobiles in cluster $c$).
Several clusters can be associated to the same group, ensuring that group
members having temporarily diverging trajectories will be kept in the group for
a minimal amount of time. Each mobile $m_i^c$ is added to the group $G(c)$ if
this group is really the probable group of the considered mobile:
$PG(m_i^c)=G(c)$. In fact, the update step aims at tracking existing members of
the group and not new comers. This procedure is summarized in
algorithm~\ref{update}.

\begin{algorithm}[h]
 \label{update}
 \Input{$\{groups_{t_c-T-1}\}$, $\{mobiles_{t_c-T}\}$}
 \Output{updated $\{groups_{t_c-T}\}$}

  $\{clusters_{t_c-T}\}=MeanShift(\{mobiles_{t_c-T}\})$\;

  \For{$c \in \{clusters_{t_c-T}\}$}
  {
    \For{$m_i^c \in \{m^c\}$}
    {
      $g_i^c=PG(m_i^c)$\;
    }
    $G(c)=\operatorname{argmax}_{g \in \{g_i^c\}}|\{g_i^c|g_i^c=g\}|$\;
  }

  \For{$m_i^c \in mobiles_{t_c-T}$}
  {
     \If{$PG(m_i^c)=G(c)$}
     {
      $G(c).add(m_i^c)$\;
     }
  }
 \caption{Update of groups.}
\end{algorithm}

The \textbf{split} of groups operates naturally. When a mobile from a group has
moved away for too many frames, its probable group becomes empty and it cannot
be added to an existing group during the update step, so it splits. It may be
part of a new group in the creation step, if it gets clustered together with
other mobiles.

Two groups $g_1$ and $g_2$ can be \textbf{merged} if two mobiles, one in each
group at frame $t_c-T+k$ ($k \in [0;T-1]$), have the same son at frame $t_c-T+l$
($l \in [k+1;T-1]$), meaning that the two mobiles will merge. The oldest group
among $g_1$ and $g_2$ is kept and all mobiles of the disappearing group are
added into the remaining group.

The group \textbf{termination} step erases old groups. Mobiles that have been
detected at a largely outdated frame (\emph{e.g.} $t_c-5T$) are deleted at frame
$t_c-T$ and empty groups are erased. As a consequence, groups having no new
mobiles for $5T$ frames are erased. All existing groups, even currently empty
ones, can potentially be updated.

Finally, the output of the group tracker, which is the input of the event
detection, is a set of tracked groups (keeping a consistent id through frames)
having properties (such as the intra-objects distance) and composed of detected
physical objects at each frame.


\begin{figure}[!htb]
  \centering
    \includegraphics[width=0.25\textwidth]{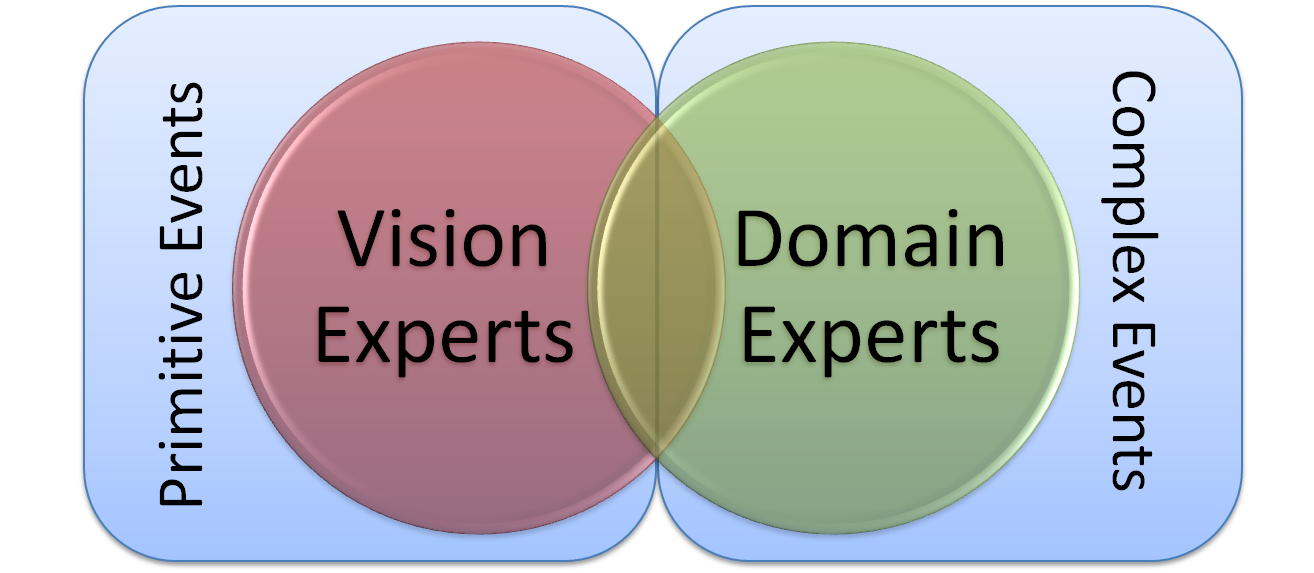}
 \caption{Knowledge modeling for video event recognition.}
\label{fig:know}
\end{figure}

\section{Event Recognition: a Generic Framework}
\label{sec:events}

In this work, a generic framework for event recognition is proposed
(\emph{ScReK}). The motivation of this framework is to use it within any video
understanding application. The genericity is obtained in terms of objects of
interest and event models.
We can identify two main parts in an event recognition process: the application
knowledge (what are the expected objects? what are the event models?) and the
event recognition algorithm.\\
Knowledge representation is a key issue for genericity. We believe that the
knowledge should be modeled with the collaboration of two different categories
of people (figure~\ref{fig:know}): vision experts (specialists in vision algorithms), and application domain
experts (specialists in the expected events of their domain). Vision experts are modeling the \textbf{objects of interest} (detected
by vision algorithms) and the \textbf{video primitives} (properties computed on the detected objects of interest). Domain experts have to model the
expected application events.\\
%
%
%
Usually, for video event recognition, knowledge is represented using OWL (Web
Ontology Language). 
Even with
tools like Prot\'eg\'e, it is difficult for a non computer specialist to
create her/his own model without a long and tedious learning of the OWL
formalism. 
The \emph{ScReK}
framework proposes its own declarative language to easily describe the application domain knowledge: the
\emph{ontology}. 
\emph{ScReK} proposes a grammar description of the objects and events using the
extended BNF (Backus Naur Form) representation. 
$O_i$, is described by its parent, $O_j$, and its attributes:
$O_i=\{a_k\}_{k=0,...O_i^n}$. The objects are defined using an inheritance
mechanism. The object $O_i$ inherits all the attributes of its parent $O_j$. The
attributes are described with the help of basic types. 11 basic types are
predefined: boolean, integer, double, timestamp, time interval, 2D point
(integer and double), 3D point (integer and double), and list of 3D points. The
user can contribute by adding new basic types. Moreover, a history of the values
of a basic type is automatically stored. It is useful for vision primitives
based on the evolution of a value in time (\emph{e.g.} trajectory).\\
For group behavior recognition, detected group objects within the video
sequence and scene context objects (zone, equipment) are described. The scene
context objects help to recognize specific events (\emph{e.g.} by defining
a forbidden access zone or a threshold). For instance, the class of group
objects is defined as follows in the ScReK language:

{\footnotesize 
\begin{lstlisting}
class Group:Mobile {
   const false;
   CSInt NumberOfMobiles;
   CSDouble AverageDistMobiles;}
\end{lstlisting}
}


%
A \textit{Group} is a \textit{Mobile} and it inherits all the attributes of a
\textit{Mobile} object (3D size, 3D position,...). A \textit{Group} is not
constant (dynamic, \emph{i.e.} its attributes values can change throughout
time). One of its attributes, \textit{NumberOfMobiles} is the number of objects
which compose the group.\\ 
The second kind of knowledge to represent is the event models. They are
composed of 6 parts: (1) the \textbf{type} of the \textbf{scenario} can be one of the following:
  \emph{PrimitiveState}, \emph{CompositeState}, \emph{PrimitiveEvent},
  \emph{CompositeEvent}, from the simplest to the most complex events. (2) the \textbf{name} of the event model which can be referenced for more complex events. (3) the list of \textbf{physical objects} (\emph{i.e.} objects of interest)
  involved in the event. The type of the objects is depending on the application
  domain. (4) the list of \textbf{components} contains the sub-events composing the event model. (5) the list of \textbf{constraints} for the physical objects or the
  components. The constraints can be temporal (between the components) or
  symbolic (for physical objects). (6) the \textbf{alarm} information describes the importance of the scenario
  model in terms of urgency. Three values are possible, from less urgent to more
  urgent: {\small \verb+NOTURGENT+}, {\small \verb+URGENT+}, {\small
  \verb+VERYURGENT+}. The alarm level can be used
  to filter recognized events, for displaying only important events to the
  user. Hereafter is a sample event model:

{\footnotesize 
\begin{lstlisting}
CompositeEvent(browsing,
  PhysicalObjects((g:Group),(e:Equipment))
  Components((c1:Group_Stop(g))
  	(c2:Group_Near_Equipment(g,e)))
  Constraints((e->Name = "shop_window"))
  Alarm ((Level : URGENT)))
\end{lstlisting}
}

The application domain expert models the event \textit{browsing} by ``a group is stopped in
front of the shop-window" with the model above. The vision expert models the sub-event
\textit{Group\_Near\_Equipment} (by measuring the distance between a group and
an equipment) and \textit{Group\_Stop} (by computing the speed of a group).\\

The last part of the event recognition framework is the recognition algorithm
itself. The proposed algorithm solves spatio-temporal constraints on the
detected groups. The usual algorithms to recognize such events can be time
consuming. The
\emph{ScReK} framework proposes 
to define optimal event models: at most two components, at most one temporal
constraint (Allen's algebra) between these components. This property is not
restrictive since all event models can be optimized in this format. Thanks to
the optimal property, the event model tree is computed. The tree defines which
sub-event (component) triggers the recognition of which event: the sub-event
which happens last in time triggers the recognition of the global event. For
instance, the event A has two components B and C with constraint: B
\textit{before} C. The recognition of C triggers the recognition of A. The tree
triggers the recognition of the only events that can happen, decreasing the
computation time.\\
%
%
%
The first step of the event recognition process is to recognize all the possible
simple events (most of these events are based on the vision primitives) by
instantiating all the models with the detected objects (\emph{e.g.}
instantiating the model \verb?Group_Stays_Inside_Zone? (takes as input one group
and one zone) for all the detected groups and all the zones of the context). The
second step consists in recognizing complex events according to the event model
tree and the simple events previously recognized. The final step checks if the
recognized event at time $t$ has been already recognized previously to update
the event (end time) or create a new one.

 \begin{figure}[!htb]
  \centering
     \includegraphics[width=0.45\textwidth]{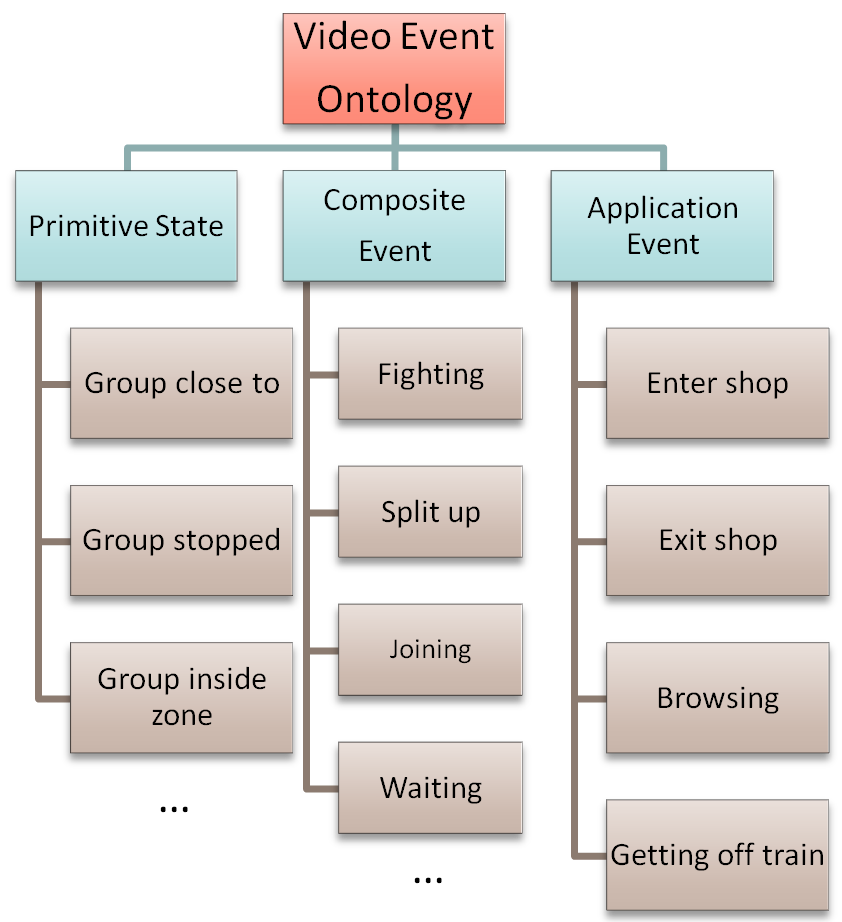}
  \caption{Proposed group event ontology.}
 \label{fig:ontology}
 \end{figure}

\section{Results}
\label{sec:res}


People detection is an input to group detection. We compared several methods to
validate our choice of method. Table~\ref{tb:resPeopleDetectionFiltered} sums up
the results of an evaluation done on a 36006 frames sequence (approximately 2
hours of video) in which 37 ground truth (GT) objects (people) have been
annotated. \cite{Corvee2011} is a feature-based people detector
whereas~\cite{Nghiem2009} and~\cite{Yao2007} both perform motion segmentation
and classification of detected objects. The method C combines the first two
methods for a more robust detection than each one separately. The method
from~\cite{Yao2007} gives the best results and is used as input of the group
tracking process. This method learns a background model, resulting in better
motion segmentation and better detection of small objects (far from the camera)
and static objects. The drawback is the time necessary to learn the model and
the low speed of the background-subtraction.

\rowcolors{1}{Gainsboro}{White}

\begin{small}

\begin{table}
    \centering
   \begin{tabular}{|c|c|c|c|c|}
        \cline{2-5}
	\hiderowcolors
        \multicolumn{1}{c|}{} & \cite{Corvee2011} & \cite{Nghiem2009} & C &
\cite{Yao2007} \\
	\hline\\[-1.2em]\hline
	\showrowcolors
        True Positives (TP) & 3699 & 3897 & 4547 & 6559 \\ \hline
        False Positives (FP) & 1379 & 185 & 125 & 128 \\ \hline
        False Negatives (FN) & 3572 & 3374 & 2724 & 2598 \\ \hline
        Precision (global) & 0.73 & 0.95 & 0.97 & 0.98 \\ \hline
        Sensitivity (global) & 0.51 & 0.54 & 0.62 & 0.72 \\ \hline
    \end{tabular}
    \caption{Comparison several people detection methods.}
      \label{tb:resPeopleDetectionFiltered}
\end{table}
\end{small}





 \begin{figure}[!htb]
  \centering
     \includegraphics[width=0.46\textwidth]{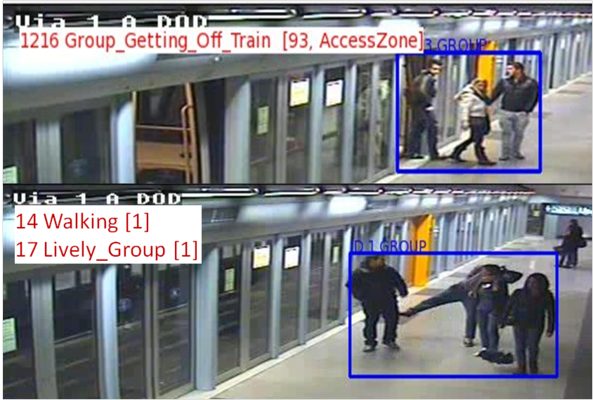}
  \caption{Event detection in Turin subway.}
 \label{fig:vanaheim}
 \end{figure}

We have performed evaluation of the group tracking algorithm using 4 different
views from 3 datasets: videos recorded for the european project
VANAHEIM\footnote{\url{https://www.vanaheim-project.eu/}} in the Turin subway
(figure~\ref{fig:vanaheim}), videos recorded for the european project
ViCoMo\footnote{\url{http://www.vicomo.org/}} at the Eindhoven airport
(figure~\ref{fig:eindhoven}) and videos from the benchmarking
CAVIAR\footnote{\url{http://homepages.inf.ed.ac.uk/rbf/CAVIARDATA1/} } dataset:
the INRIA entrance and the shopping center corridor. In
tables~\ref{tb:ResultsCAVIAR} and~\ref{tb:Results} the following metrics are
used. The \emph{fragmentation} metric computes throughout time how many tracked
objects are associated with one reference object (ground truth data). The
\emph{tracking time} metric measures the percentage of time during which a
reference object is correctly tracked. The \emph{purity} computes the number of
reference object IDs per tracked object. A value close to 1 is significative of
a good tracking.

Table~\ref{tb:ResultsCAVIAR} shows group detection and tracking results on 16
sequences from the CAVIAR dataset. 
The first 9 are sequences from INRIA, and the remaining are from the shopping
center corridor. One can notice that the shopping view is far more challenging
than the hall because more people are visible and there are more occlusions due
to the low position of the camera. Table~\ref{tb:Results} contains the results
of this evaluation on 3 annotated sequences (resp. 128, 1373 and 17992 frames)
from the Turin subway dataset. In both tables, detection results are good for
almost all sequences. In the sequence c2ls1, ground truth groups in the far end
of the corridor fail to be detected because of the limitations of the
background-subtraction method. Tracking shows good results with a few
exceptions. For instance, sequence 2 of table~\ref{tb:Results} contains a main
group present in the foreground for the whole duration of the sequence. This
group is correctly tracked with only one id-switch, but many groups are
annotated far in the background and are difficult to detect for the motion
segmentation algorithm. Their sparse detection results in many id-switches for
group tracking. At the best of our knowledge, there is no possibility of
comparing our method to an existing one (no public results or code available).

\begin{footnotesize}

\setlength{\tabcolsep}{0.01cm}

\begin{table}
    \centering
   \begin{tabularx}{\linewidth}{|p{0.85cm}|X|X|X|X|X||X|X|X|}
	\hhline{~--------}
	\hiderowcolors
	\multicolumn{1}{c|}{} &
\multicolumn{5}{{>{\columncolor{Linen}}c||}}{Detection} &
\multicolumn{3}{{>{\columncolor{Linen}}c|}}{Tracking}\\
        \cline{1-9}
        \begin{sideways}{Seq}\end{sideways} &
	\begin{sideways}{TP}\end{sideways} &
	\begin{sideways}{FP}\end{sideways} &
	\begin{sideways}{FN}\end{sideways} &
	\begin{sideways}{Prec}\end{sideways} &
	\begin{sideways}{Sens}\end{sideways} &
	\begin{sideways}Frag\end{sideways} &
	\begin{sideways}TT\end{sideways} &
	\begin{sideways}Purity\end{sideways} \\
\hlinewd{1pt}
	\showrowcolors
	fc & 125 & 101 & 3 &
	0.55 & 0.98 & 1 & 0.97 & 1 \\
	\hline
	fomd & 159 & 0 & 8 &
	1 & 0.95 & 1 & 0.95 & 1 \\
	\hline
	fra1 & 139 & 0 & 67 &
	1 & 0.67 & 1 & 0.61 & 1 \\
	\hline
	fra2 & 141 & 0 & 55 &
	1 & 0.72 & 1 & 0.70 & 1 \\
	\hline
	mc1 & 231 & 0 & 82 &
	1 & 0.74 & 1 & 0.72 & 1 \\
	\hline
	ms3g & 145 & 37 & 37 &
	0.80 & 0.80 & 1 & 0.61 & 1 \\
	\hline
	mwt1 & 156 & 0 & 89 &
	1 & 0.64 & 1 & 0.28 & 1 \\
	\hline
	mwt2 & 336 & 0 & 268 &
	1 & 0.56 & 1 & 0.53 & 1 \\
	\hline
	sp & 165 & 4 & 36 &
	1 & 0.82 & 1 & 0.67 & 1 \\
	\hlinewd{1pt}
        c2es1 & 858 & 652 & 487 &
	0.57 & 0.64 & 0.58 & 0.41 & 0.81 \\
	\hline
        c2es3 & 1093 & 550 & 735 &
	0.66 & 0.60 & 0.66 & 0.38 & 0.83 \\
	\hline
        c2ls1 & 788 & 1664 & 655 &
	0.32 & 0.55 & 0.50 & 0.13 & 1 \\
	\hline
	c3ps1 & 1298 & 1135 & 210 &
	0.54 & 0.86 & 1 & 0.68 & 1 \\
	\hline
	cosme2 & 1119 & 852 & 35 &
	0.57 & 0.97 & 0.25 & 0.60 & 1 \\
	\hline
	csa1 & 269 & 163 & 0 &
	0.63 & 1 & 1 & 0.96 & 1 \\
	\hline
	cwbs1 & 2224 & 89 & 1090 &
	0.96 & 0.67 & 1 & 0.45 & 0.80 \\
	\hline
    \end{tabularx}

    \caption{Results of group detection and tracking on 16 CAVIAR sequences.
(Seq -- official sequence name, Prec -- Precision, Sens -- Sensitivity, Frag --
Fragmentation, TT -- Tracking Time)}
      \label{tb:ResultsCAVIAR}
\end{table}

\end{footnotesize}


\rowcolors{1}{Gainsboro}{White}

\begin{small}

\begin{table}
    \centering

   \begin{tabularx}{\linewidth}{|X|X|X|X|X|X||X|X|X|}
	\hhline{~--------}
	\hiderowcolors
	\multicolumn{1}{c|}{} &
\multicolumn{5}{{>{\columncolor{Linen}}c||}}{Detection} &
\multicolumn{3}{{>{\columncolor{Linen}}c|}}{Tracking}\\
        \cline{1-9}
        \begin{sideways}{Seq}\end{sideways} &
	\begin{sideways}{TP}\end{sideways} &
	\begin{sideways}{FP}\end{sideways} &
	\begin{sideways}{FN}\end{sideways} &
	\begin{sideways}{Prec}\end{sideways} &
	\begin{sideways}{Sens}\end{sideways} &
	\begin{sideways}Frag\end{sideways} &
	\begin{sideways}TT\end{sideways} &
	\begin{sideways}Purity\end{sideways} \\
\hlinewd{1pt}
	\showrowcolors
        1 & 65 & 0 & 6 &
	1 & 0.91 & 1 & 0.86 & 1 \\
	\hline
        2 & 1346 & 69 & 318 &
	0.95 & 0.80 & 0.80 & 0.14 & 0.91 \\
	\hline
        3 & 6977 & 1677 & 4594 &
	0.80 & 0.60 & 0.40 & 0.32 & 0.98 \\
	\hline
    \end{tabularx}

    \caption{Results of group detection and tracking on 3 sequences from the
Turin subway. (Prec -- Precision, Sens -- Sensitivity, Frag -- Fragmentation, TT
-- Tracking Time)}
      \label{tb:Results}
\end{table}

\end{small}

\begin{table}
\begin{center}
\begin{tabular}{ | l | c || c | c | c | }
   \hhline{~----}
   \multicolumn{1}{c|}{} & GT & TP & FP & FN \\
   \hline 
   fighting & 2 & 1 & 0 & 1 \\
   \hline
   split up & 3 & 3 & 0 & 0 \\
   \hline
   joining & 3 & 3 & 0 & 0 \\
   \hline
   shop enter & 5 & 5 & 0 & 0 \\
   \hline
   shop exit & 6 & 6 & 1 & 0 \\
   \hline
   browsing & 3 & 3 & 1 & 0 \\
   \hline
   getting off train & 10 & 9 & 8 & 1 \\
   \hline
 \end{tabular}
  \caption{Group event recognition for the 3 video datasets}
      \label{tb:ResultsEvents}
\end{center}
\end{table}


One major achievement of this paper is an ontology for group events based on video sensor (figure \ref{fig:ontology}). The ontology is composed
of 49 event models (45 models are generic and re-usable in any application with
groups (\textit{Group\_stop, Group\_lively,...}), 4 models are specifically
defined for the applications of this paper (the events depend on the application
context, \textit{enter shop},...)). The events have been modeled with help of metro surveillance staff.\\



\begin{figure}[ht!]
 \centering
 \includegraphics[width=0.23\textwidth]{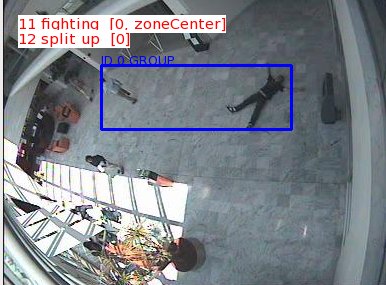}
 \includegraphics[width=0.23\textwidth]{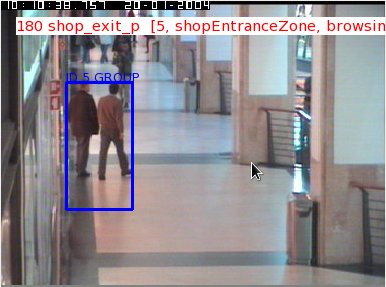}\\
 \includegraphics[width=0.23\textwidth]{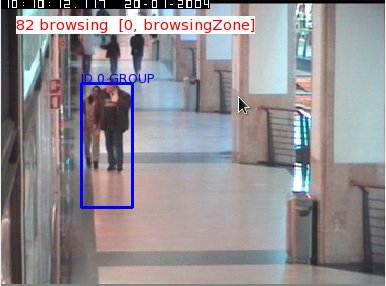}
 \includegraphics[width=0.23\textwidth]{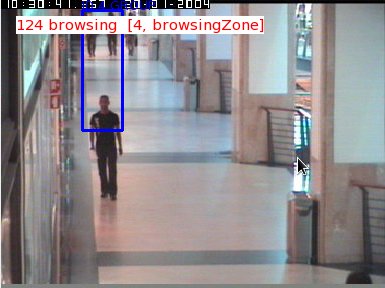}
 \caption{Group event recognition for the CAVIAR sequences. a. Fighting then splitting. b. Exit from the shop. c. Browsing. d.
The mis-detected group (ghost due to reflections) is browsing.}
 \label{fig:fight}
\end{figure}

The results of the group event recognition are given in
table~\ref{tb:ResultsEvents} for the interesting events. Examples of event
recognition are shown in figures~\ref{fig:eindhoven},~\ref{fig:vanaheim}
and~\ref{fig:fight}. There is only a few instances of each event because we
only focus on meaningful group events. The events are correctly recognized with
low false positive and false negative rates. Most of the false positive
detections for the event \textit{getting off train} are due to the fact that the
door in the foreground is detected as a person when open. The errors can be
corrected by adding a new video primitive: \textit{door detector}.






\section{Conclusions}


We propose a generic, plug and play framework for event recognition from videos:
\emph{ScReK}. The scientific community can share a common ontology composed of
event models and vision primitives. We demonstrate this framework on 4 group
behavior recognition applications, using a novel group tracking approach. This
approach gives satisfying results even on very challenging datasets (numerous
occlusions and long duration sequences) such as in figure~\ref{fig:vanaheim}.
The vision primitives are based on global attributes of groups (position, speed,
size). The proposed event detection approach correctly recognizes events but
shows its limitation for some specific events (\emph{e.g.} fighting is best
characterized by internal group movement). Adapted vision primitives, such as
optical flow, solve specific limitations and are easy to plug into \emph{ScReK}.
Moreover, in this work the gap between video data and semantical events is
modeled manually by vision experts, the next step is to learn automatically the
vision primitives.\\


\textbf{Acknowledgment} This work was supported partly by the Video-Id, ViCoMo,
Vanaheim, and Support projects. However, the views and opinions expressed herein
do not necessarily reflect those of the financing institutions.

{\small
\bibliographystyle{ieee}
\bibliography{biblio}
}

\end{document}